\documentclass[letterpaper, 10 pt, conference]{ieeeconf}  

\IEEEoverridecommandlockouts                              

\usepackage{amsfonts}
\usepackage{bm}
\usepackage{booktabs}
\usepackage{graphicx}
\usepackage{mathtools}
\usepackage{subcaption}

\makeatletter
\let\NAT@parse\undefined
\makeatother
\usepackage[pagebackref=false,breaklinks=true,letterpaper=true,colorlinks,bookmarks=false]{hyperref}

\newcommand{\stufig}[5]                                       
{
	\begin{figure}[#5]
		\begin{center}
			\includegraphics[#1]{#2}
			\caption{#3}
			\label{#4}
		\end{center}
		\vspace{-1.5\baselineskip}
	\end{figure}
}

\newcommand{\stufigstar}[5]                                   
{
	\begin{figure*}[#5]
		\begin{center}
			\includegraphics[#1]{#2}
			\caption{#3}
			\label{#4}
		\end{center}
		\vspace{-1.5\baselineskip}
	\end{figure*}
}

\newenvironment{stusubfig}[1]
{
	\begin{figure}[#1]
		\begin{center}
		}
		{
		\end{center}
	\end{figure}
}

\newenvironment{stusubfig*}[1]
{
	\begin{figure*}[#1]
		\begin{center}
		}
		{
		\end{center}
	\end{figure*}
}

\title{\LARGE \bf
Real-Time Hybrid Mapping of Populated Indoor Scenes \\ using a Low-Cost Monocular UAV
}

\author{
	\begin{tabular}{c}
		\begin{tabular}{c@{\hskip 0.85cm}c@{\hskip 0.85cm}c@{\hskip 0.85cm}c}
			Stuart Golodetz & Madhu Vankadari$^*$ & Aluna Everitt$^*$ & Sangyun Shin$^*$ \\
			& Andrew Markham & Niki Trigoni \\
		\end{tabular}
	\end{tabular}%
	\thanks{All authors are with the University of Oxford.}%
	\thanks{M.\ Vankadari, A.\ Everitt and S.\ Shin assert joint second authorship.}%
	\thanks{E-mail: \{stuart.golodetz, madhu.vankadari, aluna.everitt, sangyun.shin, andrew.markham, niki.trigoni\}@cs.ox.ac.uk.}%
}

\begin{document}

\maketitle
\thispagestyle{empty}
\pagestyle{empty}

\begin{abstract}
Unmanned aerial vehicles (UAVs) have been used for many applications in recent years, from urban search and rescue, to agricultural surveying, to autonomous underground mine exploration. However, deploying UAVs in tight, indoor spaces, especially close to humans, remains a challenge. One solution, when limited payload is required, is to use micro-UAVs, which pose less risk to humans and typically cost less to replace after a crash. However, micro-UAVs can only carry a limited sensor suite, e.g.\ a monocular camera instead of a stereo pair or LiDAR, complicating tasks like dense mapping and markerless multi-person 3D human pose estimation, which are needed to operate in tight environments around people. Monocular approaches to such tasks exist, and dense monocular mapping approaches have been successfully deployed for UAV applications. However, despite many recent works on both marker-based and markerless multi-UAV single-person motion capture, markerless single-camera multi-person 3D human pose estimation remains a much earlier-stage technology, and we are not aware of existing attempts to deploy it in an aerial context. In this paper, we present what is thus, to our knowledge, the first system to perform simultaneous mapping and multi-person 3D human pose estimation from a monocular camera mounted on a single UAV. In particular, we show how to loosely couple state-of-the-art monocular depth estimation and monocular 3D human pose estimation approaches to reconstruct a hybrid map of a populated indoor scene in real time. We validate our component-level design choices via extensive experiments on the large-scale ScanNet and GTA-IM datasets. To evaluate our system-level performance, we also construct a new \emph{Oxford Hybrid Mapping} dataset of populated indoor scenes.
\end{abstract}

\section{Introduction}

Recent years have seen huge improvements in the flight stability and obstacle avoidance capabilities of unmanned aerial vehicles, driven by applications including aerial search and rescue \cite{Waharte2010}, aerial tracking and surveillance \cite{Du2018}, drone cinematography \cite{Mademlis2018}, robotic agriculture \cite{Kim2019}, and the exploration of everything from mines \cite{Papachristos2019} to other planets \cite{MarsHelicopter}. However, deploying drones in confined indoor spaces close to people remains challenging. This is unfortunate, because numerous applications, from awareness systems for emergency responders to indoor drone cinematography for film-makers, could benefit significantly from such a capability.

\begin{stusubfig}{!t}
	\begin{subfigure}{.31\linewidth}
		\centering
		\includegraphics[width=\linewidth]{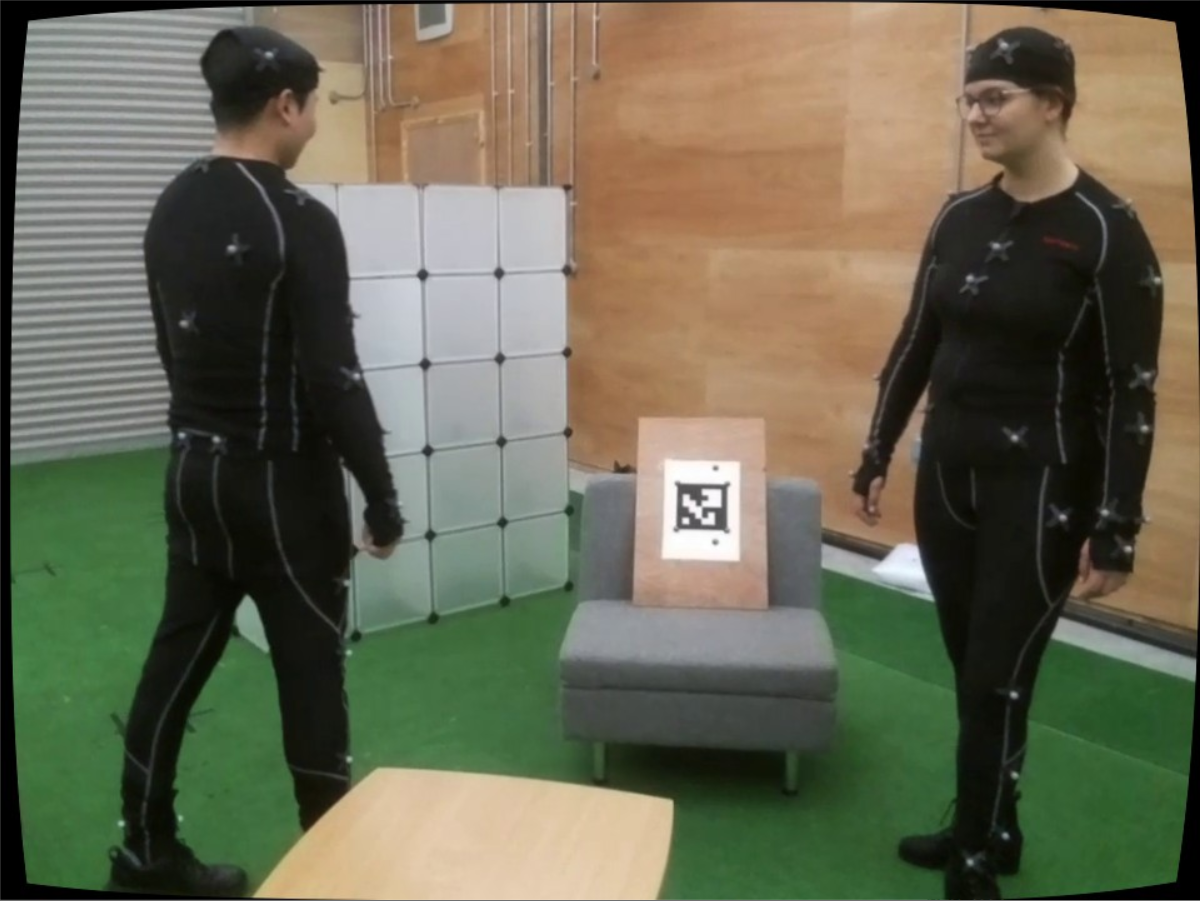}
	\end{subfigure}%
	\hspace{2mm}%
	\begin{subfigure}{.31\linewidth}
		\centering
		\includegraphics[width=\linewidth]{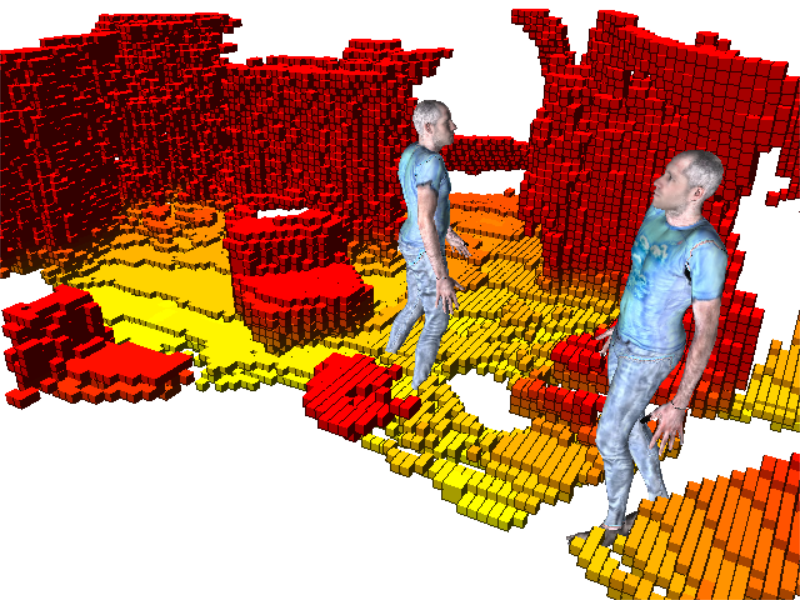}
	\end{subfigure}%
	\hspace{2mm}%
	\begin{subfigure}{.31\linewidth}
		\centering
		\includegraphics[width=\linewidth]{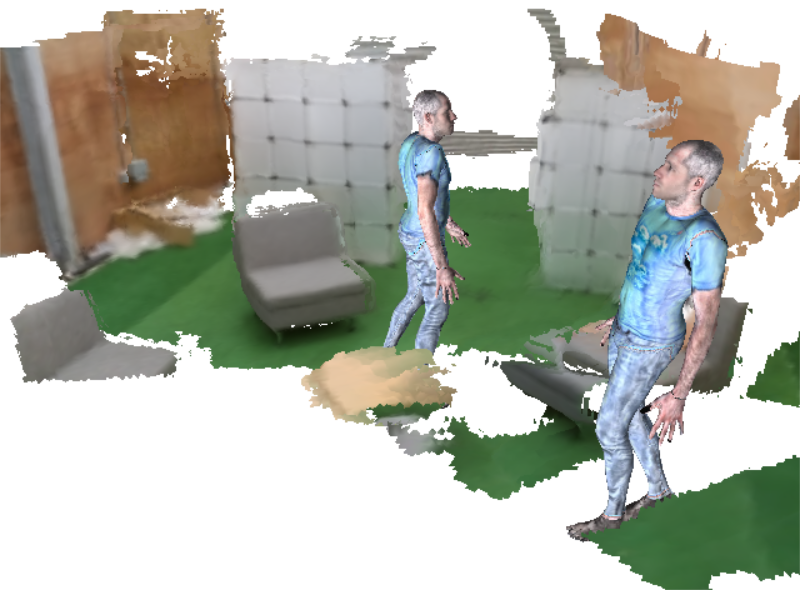}
	\end{subfigure}%
	\caption{Reconstructing hybrid maps of populated indoor scenes from a monocular camera mounted on a single UAV. We render from free view to better show the map.}
	\label{fig:teaser}
	\vspace{-2\baselineskip}
\end{stusubfig}

To operate in such an environment, it is helpful for a drone to be able to both map its geometry and detect/track the people moving within it, ideally in real time. At the same time, however, the physical constraints imposed by the environment encourage the use of a small drone (e.g.\ $\approx$10cm in diameter) that can more easily avoid obstacles/people, fit through narrow openings, and pose less risk in a collision. Unfortunately, most existing drones of that size are equipped with only a monocular camera, rather than the depth sensors, stereo cameras or LiDAR that can be deployed on large drones. This setup, whilst ubiquitous in modern consumer-grade drones and mobile phones, complicates the mapping and person detection/tracking tasks, since both monocular depth estimation (which underlies most monocular mapping approaches) and monocular 3D human pose estimation are ill-posed problems. Nevertheless, numerous approaches to both tasks do exist: typically, these use multiple images captured over time and/or leverage prior knowledge.

For monocular depth estimation (\S\ref{subsec:related-monoreconstruction}), many real-time methods exist. However, for monocular 3D human pose estimation (\S\ref{subsec:related-monohumanpose}), very few methods can estimate the skeletons of multiple people in real time. Moreover, whilst single-person 3D human pose estimation has been performed from a monocular drone, many approaches rely on markers \cite{Nageli2018} or offline post-processing \cite{Zhou2018,Saini2019,Ho2021}. To our knowledge, no approach has yet deployed real-time markerless multi-person 3D human pose estimation for a monocular drone. Notably, simultaneous scene reconstruction and 3D human pose estimation from a drone (monocular or not) has also received limited attention. Some cinematography (\S\ref{subsec:related-aerialmocap}) works have used an offline map to allow drones to avoid obstacles during filming \cite{Ho2021}, or performed online mapping and (non-skeletal) person tracking using a large LiDAR-equipped drone \cite{Bonatti2019}. However, we are not aware of any works that currently perform both online mapping and skeletal pose estimation, particularly for multiple people and from a monocular drone.

In this paper, we propose what is thus, to our knowledge, the first system to perform simultaneous mapping and multi-person 3D human pose estimation from a monocular camera mounted on a single UAV. In particular, we investigate how to loosely couple state-of-the-art monocular depth estimation and 3D human pose estimation approaches to reconstruct a hybrid map of a populated scene in real time. Our approach is portable, requiring only a small, stable, low-cost monocular UAV (e.g.\ a DJI Tello) and a GPU-enabled laptop, and can be deployed in tight, indoor spaces close to people.

To validate our component-level design choices, we conduct experiments on ScanNet \cite{Dai2017CVPR} and GTA-IM \cite{Cao2020}, which (respectively) target 3D reconstruction and indoor motion / human-scene interaction (\S\ref{subsec:experiments-scannet} and \S\ref{subsec:experiments-gta-im}). To evaluate our system's overall performance, we also collect a new \emph{Oxford Hybrid Mapping} dataset of populated indoor scenes (\S\ref{subsec:experiments-ohm}). Our dataset and source code will be made available online.

\section{Related Work}
\label{sec:relatedwork}

\subsection{Real-Time Dense Monocular 3D Reconstruction}
\label{subsec:related-monoreconstruction}

Dense monocular 3D reconstruction methods can be divided into two types: depth-based methods, which estimate dense depth images and fuse them to reconstruct the scene, and representation-based methods (e.g.\ \cite{Sun2021}), which predict all/part of a scene representation. Since our system needs to output multiple scene representations, and representation-based methods typically target only one representation type (e.g.\ a TSDF \cite{Curless1996}), we restrict our attentions to real-time depth-based methods, since they are a better fit for our work.

Within depth-based methods, single-image methods like \cite{Laina2016} estimate the corresponding depth image for a single input image, and multi-image methods leverage multiple images over time to estimate depth. State-of-the-art multi-image methods tend to be more accurate than single-image ones, owing to the greater amount of information they have to make their predictions, and so we further restrict our focus to real-time depth-based multi-image methods in this paper.

\emph{Multi-view stereo} (MVS) methods \cite{Pradeep2013,Greene2016} establish correspondences between pixels in different images and triangulate matching pixels to determine their depths. For example, MVDepthNet \cite{Wang20183DV} uses an encoder-decoder with skip connections to estimate inverse depth from a cost volume. MultiViewStereoNet \cite{Greene2021} compensates for known viewpoint changes when extracting features, reducing the need for the network to learn rotation/scale invariant features.

\emph{ConvLSTM-based} methods \cite{Tananaev2018,Zhang2019} exploit ConvLSTMs to predict temporally consistent depth. They produce visually appealing results, but cannot ensure geometric correctness because their initial predictions / spatial features derive from single images, rather than multi-image triangulation.

\emph{Hybrid} methods \cite{Tateno2017,Laidlow2019} combine techniques to achieve good performance. For example, CNN-SLAM \cite{Tateno2017} predicts depth images for keyframes, then refines them using small-baseline multi-view stereo from surrounding non-keyframe images. DeepVideoMVS \cite{Duzceker2021} extends a cost volume-based encoder-decoder with a ConvLSTM cell at the bottleneck layer to leverage past scene geometry to improve depth prediction at the current time step. Unlike standard ConvLSTM methods, it can make geometrically correct predictions because of its underlying use of MVS at each time step.

\subsection{Real-Time Multi-Person Monocular 3D Pose Estimation}
\label{subsec:related-monohumanpose}

Human pose estimation methods can be either \emph{top-down} (detect people in the image, then estimate their skeletons), or \emph{bottom-up} (detect individual keypoints, then group the keypoints for different people and assemble them into skeletons).

The first near-real-time monocular multi-person approach, LCR-Net \cite{Rogez2017CVPR}, was top-down, first using a region proposal network to predict candidate regions for people in the image, then predicting a pose refinement for each of a fixed set of anchor poses and optionally choosing an anchor pose for each region. LCR-Net++ \cite{Rogez2020PAMI} later used synthetic training data and an improved backbone to improve the accuracy.

Separately, VNect \cite{Mehta2017TOG}, a real-time, single-person, bottom-up approach, predicted (for each frame) a 2D heatmap and three 3D location maps per joint. Mehta et al.\ \cite{Mehta20183DV} extended this to multiple people, replacing the location maps
with occlusion-robust pose maps and part affinity fields. Mehta et al.\ \cite{Mehta2020TOG} later proposed XNect, which uses a three-stage pipeline to achieve superior performance to \cite{Mehta20183DV}.

Two other real-time multi-person methods have been proposed. MP3D \cite{Dabral2019} is quasi top-down and appends an hourglass module to the keypoint head of Mask-RCNN \cite{He2017} to improve the heat maps. PandaNet \cite{Benzine2020} works bottom-up to estimate the skeletons of up to 60 people in under 140ms.

\subsection{Aerial Human Motion Capture}
\label{subsec:related-aerialmocap}

There has been much recent interest in drone-based 3D skeleton estimation. For example, Flycon \cite{Nageli2018} uses two monocular drones to fly autonomously around a subject wearing active markers and estimates the drones' poses and the subject's 3D skeleton in real time. Zhou et al.\ \cite{Zhou2018} use a monocular drone to autonomously fly around a subject who is moving on the spot and record a video that is post-processed offline to recover a 3D skeleton sequence. AirCap \cite{Saini2019} uses several monocular drones to autonomously fly around a markerless subject outdoors and capture images, poses and regions of interest for later offline body fitting.

\begin{figure*}[!t]
	\vspace{2mm}
	\begin{center}
		\includegraphics[width=.83\linewidth]{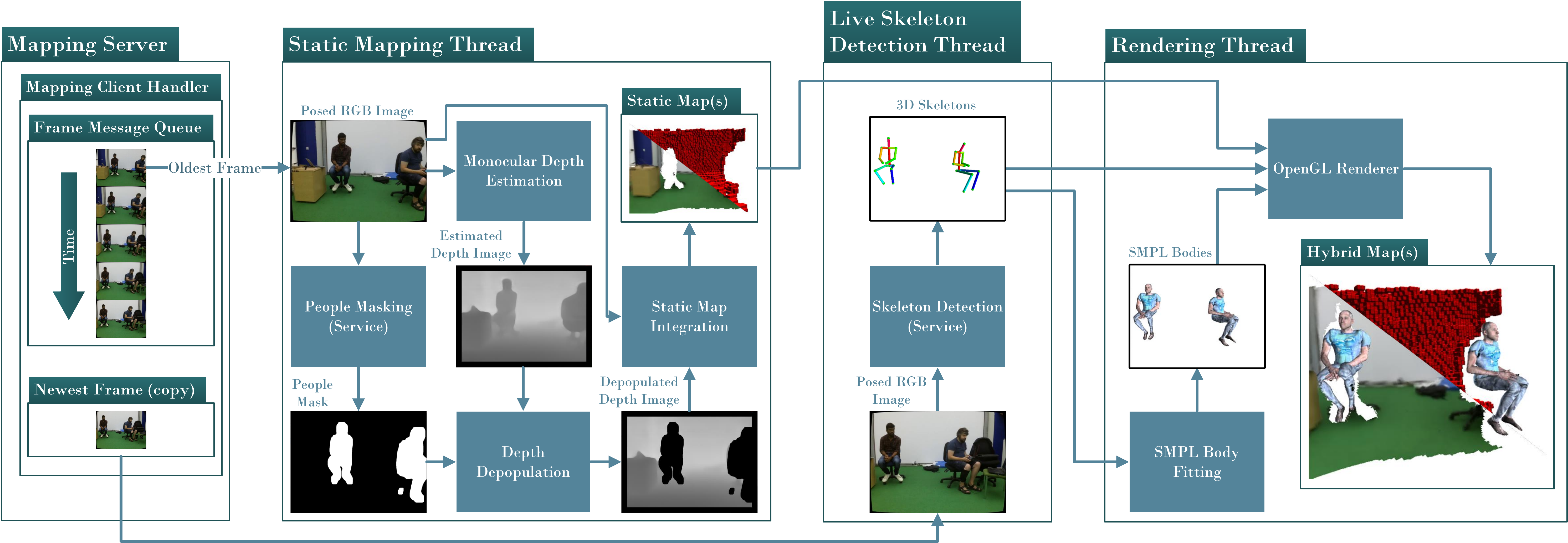}
		\caption{Our system. The client feeds RGB frames and their poses to the mapping server. Each new frame is (i) pushed onto a fixed-size queue, replacing a random frame if the queue is full, and (ii) copied for the live skeleton detector. The \emph{static mapping} thread repeatedly consumes the oldest frame from the queue and estimates a depth image for it, whilst a \emph{people masking} service produces a mask of all the people in the frame. The mask is used to depopulate the depth image, which is then fused into one or more static map representations. In parallel, the \emph{live skeleton detection} thread detects 3D skeletons in the most recent frame received from the client. The \emph{rendering} thread optionally fits SMPL \cite{Loper2015} bodies (with textures from SURREAL \cite{Varol2017}) to the skeletons (or keeps the skeletons as they are), and renders the hybrid map using OpenGL.}
		\label{fig:pipeline}
	\end{center}
	\vspace{-1.5\baselineskip}
\end{figure*}

Bonatti et al.\ \cite{Bonatti2019} use a DJI M210 equipped with a camera and a LiDAR to autonomously follow a single person, mapping the scene online using the LiDAR whilst avoiding obstacles. The person is localised, but their 3D skeleton isn't estimated.
In very recent work, Ho et al.~\cite{Ho2021} use two DJI M210s to autonomously fly around a subject who is jogging or playing football, in the context of a known 3D map. Collision avoidance is performed online, with the important limitation of requiring the scene to be mapped ahead of time, but skeletal reconstruction is deferred to a later, offline stage. Notably, whilst many of these works are extremely impressive, none of them currently attempts to perform both online mapping and markerless multi-person 3D human pose estimation from a single monocular drone.

\subsection{Human-Scene Interaction}

Various works have learnt representations of where/how humans interact with scenes and objects \cite{Savva2016,Hassan2019,Zhang2020,Hassan2021}, and used them to improve 3D human pose estimation given knowledge of the context, or to place people in plausible positions in known scenes. PiGraphs \cite{Savva2016} and PROX \cite{Hassan2019} propose datasets of people interacting with scenes. However, both use static cameras at test time and are thus unsuitable for evaluating our system. We thus performed our component-level experiments on GTA-IM \cite{Cao2020}, and collected our own dataset to evaluate our overall system (\S\ref{subsec:experiments-ohm}).

\section{Hybrid Mapping System}
\label{sec:system}

Figure~\ref{fig:pipeline} shows an overview of our system. We adopt a client-server model similar to \cite{Golodetz2018TVCG}. Our server runs on a GPU-enabled laptop and is responsible for reconstructing the hybrid map. Various clients are possible, each of which is responsible for camera tracking, and for streaming RGB images and their scale-correct 6D poses across to the server.

\subsection{Mapping Server}

A frame message queue is maintained, based on the \emph{pooled queue} data structure described in \cite{Golodetz2018TVCG}. Incoming frames from the client are stored in the queue. The \emph{static mapping} thread consumes frames from the queue, starting with the oldest, and uses them to reconstruct an Octomap \cite{Hornung2013AR} and/or a TSDF \cite{Curless1996} of the static scene. A \emph{people masker} is used to depopulate (i.e.\ remove the people from) the depth image estimated for each frame before it is fused into the static map (\S\ref{subsubsec:peoplemasking}). We also maintain a copy of the most recent frame received from the client for the \emph{live skeleton detection} thread, which detects the skeletons of any people in the scene using a multi-person monocular 3D skeleton detector such as LCR-Net \cite{Rogez2017CVPR} or XNect \cite{Mehta2020TOG}. The \emph{rendering} thread renders the static map(s) and the skeletons of the people in the scene at interactive rates using OpenGL. Optionally, SMPL \cite{Loper2015} body models can be fitted to the skeletons (\S\ref{subsubsec:bodyfitting}). These have two uses: first, they can be rendered in place of the skeletons detected by the live detector to improve the realism of the reconstructed scene; second, SMPL body silhouettes can be rendered to produce people masks, if we use a people masker based on an intermediate skeleton detection step.

\subsection{Skeleton Detection, Body Fitting and People Masking}
\label{subsec:skeletondetection}

We implement skeleton detection as an asynchronous service, thus allowing it to be performed in parallel alongside other operations. The service is passed an RGB image and its world-space 6D pose as input, and at some later point returns a list of world-space 3D skeletons and a people mask (a binary mask of all people in the image).

\subsubsection{Skeleton Detection}

To detect the skeletons, we use one of two state-of-the-art real-time monocular 3D multi-person skeleton detectors, namely LCR-Net \cite{Rogez2017CVPR} and XNect~\cite{Mehta2020TOG} (\S\ref{subsec:related-monohumanpose}). We considered also evaluating MP3D \cite{Dabral2019} and PandaNet \cite{Benzine2020}, but unfortunately neither has available code.

\subsubsection{Body Fitting}
\label{subsubsec:bodyfitting}

\stufig{width=.95\linewidth}{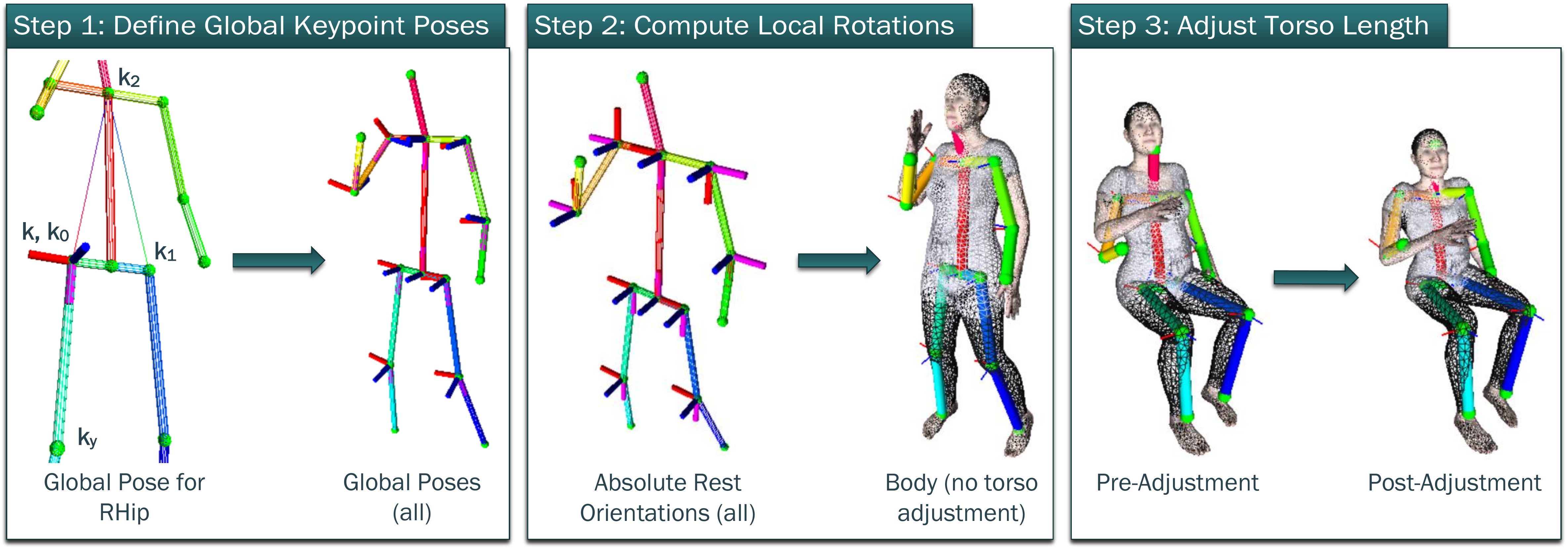}{Lifting 3D skeletons to SMPL bodies (\S\ref{subsubsec:bodyfitting}). In Step 2, we render $[{}_w^{(t)}R_m \; {}_m\hat{R}_k \; | \; \bm{t}_k^{(t)}]$ for each $k$.}{fig:smpl-fit}{!t}

To lift the detected 3D skeletons to SMPL \cite{Loper2015} body models, we propose a three-step, real-time fitting approach (see Figure~\ref{fig:smpl-fit}). This takes a skeleton whose keypoints have known positions but not known orientations, and produces a set of pose parameters $\bm{\theta}$ (the local orientations for the joints) that can be used to configure an SMPL body. Note that the skeleton may not have exactly the same set of keypoints as the body: we handle this retargeting problem by computing global orientations for only those SMPL body keypoints for which we do have a corresponding keypoint in the skeleton, and setting the local orientations of the remaining keypoints to the identity ($I_3$). Aside from the torso length ($\beta_2$), we do not currently estimate the body shape parameters $\bm{\beta}$ (see below). We found it to be sufficient for our needs in this paper to set these to neutral/zero.

Our first step defines a global pose ${}_w^{(t)}T_k = [{}_w^{(t)}R_k \; | \; \bm{t}_k^{(t)}] \in \mathbb{SE}(3)$ for each relevant keypoint $k$ at each time $t$. We associate each $k$ with (i) three distinct keypoints $k_0$, $k_1$, $k_2$ that can (indirectly) define the $z$ axis of a coordinate system whose origin is at $\bm{t}_k^{(t)}$, and (ii) a keypoint $k_y$ that can be used to define the $y$ axis of this coordinate system. Formally, we can define the coordinate axes for keypoint $k$ at time $t$ as
\begin{equation}
	\footnotesize
	\begin{aligned}
		\bm{y}^{(t)} &= \mbox{normalise}(\bm{t}_{k_y}^{(t)} - \bm{t}_k^{(t)}) \\
		\bm{n}^{(t)} &= \mbox{normalise}((\bm{t}_{k_1}^{(t)} - \bm{t}_{k_0}^{(t)}) \times (\bm{t}_{k_2}^{(t)} - \bm{t}_{k_0}^{(t)})) \\
		\bm{x}^{(t)} &= \mbox{normalise}(\bm{y}^{(t)} \times \bm{n}^{(t)}) \\
		\bm{z}^{(t)} &= \mbox{normalise}(\bm{x}^{(t)} \times \bm{y}^{(t)}).
	\end{aligned}
\end{equation}
Note that any or all of $k_0$, $k_1$, $k_2$ and $k_y$ may be occluded in a particular image. We delegate this problem to the skeleton detector, which is assumed to produce either a complete (if potentially inaccurate) skeleton or nothing at all in such cases (both of the skeleton detectors we consider work in this way).

Our second step then takes the set of global poses and the known skeletal structure (as expressed by a function $\pi$ that maps each keypoint $k$ to its parent $\pi(k)$ in the skeleton), and computes local orientations for the keypoints (with respect to their parents in the skeleton) that can be used to configure an SMPL body model. To do this, we first specify the relative rest orientation ${}_m\hat{R}_k \in \mathbb{SO}(3)$ of each keypoint with respect to the person's mid-hip $m$. We can then compute the local orientation ${}_{\pi(k)}^{(t)}\bar{R}_k \in \mathbb{SO}(3)$ of each keypoint $k$ at time $t$ as
\begin{equation}
	\footnotesize
	\begin{aligned}
		& {}_m\hat{R}_{\pi(k)} \; {}_w^{(t)}R_{\pi(k)}^{-1} \; {}_w^{(t)}R_k \; {}_m\hat{R}_k^{-1} \\
		=& {}_m\hat{R}_{\pi(k)} \; {}_{\pi(k)}^{(t)}R_w \; {}_w^{(t)}R_k \; {}_k\hat{R}_m \\
		=& {}_m\hat{R}_{\pi(k)} \; {}_{\pi(k)}^{(t)}R_k \; {}_k\hat{R}_m.
	\end{aligned}
\end{equation}
Finally, we estimate the torso length of the SMPL body ($\beta_2$) to achieve a better fit to the skeleton. To do this, we calculate the torso lengths $\ell_{\mathit{torso}}^{(s)}$ and $\ell_{\mathit{torso}}^{(b)}$ (the distances between the neck and mid-hip keypoints) in both the skeleton and the neutral SMPL body, respectively, and then define $\beta_2 = \kappa (\ell_{\mathit{torso}}^{(s)} - \ell_{\mathit{torso}}^{(b)})$, where here $\kappa$ was empirically set to $100$.

Since our main focus was on representing the locations and poses of the people in the map in real time, rather than on reconstructing their body shapes, we found this approach to be adequate for our needs, and to have the advantages of being extremely fast, easy to adapt to input skeletons with differing sets of keypoints, and straightforward to implement.

\subsubsection{People Masking}
\label{subsubsec:peoplemasking}

To generate the people mask for each frame, we consider: (i) rendering (conservative) bounding volumes fit to the detected skeletons, (ii) rendering silhouettes of the SMPL bodies fit to the skeletons (\S\ref{subsubsec:bodyfitting}), and (iii) direct instance segmentation of the RGB image. As we consider two different skeleton detectors (LCR-Net and XNect), this yields $5$ people masking variants in total: we compare their performance using GTA-IM \cite{Cao2020} (\S\ref{subsubsec:experiments-gta-im-peoplemasking}).

\emph{Bounding Volumes}. We specify (conservative) bounding volumes (cylinders and spheres) for the bones in each skeleton. We use OpenGL to render them for all detected skeletons to an off-screen frame buffer to produce the mask. This is fast ($\approx$4ms/skeleton), and the use of conservative volumes adds some robustness for bones that were inaccurately estimated by the underlying skeleton detector. However, if the detector entirely misses a person in the image, this approach will fail. Moreover, the large bounding volumes can mask out too much of the static scene in the background of an image.

\emph{Body Silhouettes}. We render the SMPL bodies, in white and unlit, to an off-screen frame buffer to create a binary people mask. This is also fast ($\approx$13ms/skeleton, including fitting), and has the advantage of producing more accurate segmentations than the bounding volumes, reducing the amount of static scene that is masked out. However, it still relies on the robustness of the underlying skeleton detector.

\emph{Instance Segmentation}. We use Mask R-CNN \cite{He2017} to produce the mask. This is comparatively slow ($\approx$200ms/frame), but we would expect it to perform better than approaches based on intermediate monocular 3D skeleton detection, owing to the relative difficulties of the two problems and the maturity levels of the two different sub-fields.

\subsection{Static Mapping}
\label{subsec:staticmapping}

As per Figure~\ref{fig:pipeline}, the static mapping thread takes an RGB image and its world-space 6D pose, and runs monocular depth estimation and people masking (\S\ref{subsubsec:peoplemasking}) on them in parallel. The people mask is used to depopulate (i.e.\ remove the people from) the estimated depth image, which is then fused into one or more static maps (in the current system, an Octomap \cite{Hornung2013AR} and/or a TSDF \cite{Curless1996}).

For monocular depth estimation, we build on two real-time multi-view stereo (MVS) approaches, MVDepthNet \cite{Wang20183DV} and DeepVideoMVS \cite{Duzceker2021} (\S\ref{subsec:related-monoreconstruction}). MVDepthNet was an early real-time-capable MVS method, whilst DeepVideoMVS is a very recent method that has achieved state-of-the-art depth estimation performance on multiple datasets.

\subsubsection{MVDepthNet}

MVDepthNet takes two overlapping views of the scene and their known poses, and estimates a depth image. To convert it into a monocular depth estimator, we implement a keyframe-based system that adds a new keyframe whenever (i) the translation to the closest keyframe is $> 5$cm, or (ii) the angle to the closest keyframe is $> 5^\circ$. To estimate the depth for each incoming frame, we first select a suitable keyframe to use with it for multi-view stereo, and then pass both it and the keyframe (and their poses) to MVDepthNet. We select the keyframe that maximises the following score (a simplified version of that in \cite{Yang2020TVCG}):
\begin{equation}
\footnotesize
S^f(k) = \mathbb{I}[\Delta_t^f(k) \ge \tau_t \mbox{ and } \Delta_\theta^f(k) \le \tau_\theta] e^{(-(\Delta_t^f(k) - \delta_t)^2 / \sigma_t^2)}
\end{equation}
In this, $\mathbb{I}$ denotes the binary indicator function, $f$ denotes the frame, $k$ denotes a keyframe, $\Delta_t^f$ and $\Delta_\theta^f$ respectively denote the baseline (m) and angle (${}^\circ$) between $f$ and $k$, $\delta_t = 0.4$, $\sigma_t = 0.2$, $\tau_t = 0.025$ and $\tau_\theta = 20$.

\subsubsection{DeepVideoMVS}

DeepVideoMVS provides a built-in keyframe-based system, and takes as input just a single image and its known pose. Full details can be found in \cite{Duzceker2021}.

\subsubsection{Post-Processing}
\label{subsubsec:postprocessing}

When constructing a map, it is typically important to achieve high levels of both accuracy and completeness. There is to some extent a trade-off between the two, and depth estimators commonly try to balance them by estimating dense depth maps whilst targeting accuracy through their loss functions. However, even with state-of-the-art methods, it is hard to predict an accurate depth for every pixel in an image, especially in textureless / partially occluded regions. There is thus a chance to achieve improved accuracy without significantly affecting \emph{reconstruction} (as opposed to depth map) completeness by finding and filtering out such pixels. In our system, we thus perform a sequence of post-processing steps on the estimated depth images:

\emph{Temporal Inconsistency Filtering:} First, we filter out any pixel in the current depth image $D_t$ whose estimated depth is inconsistent with the previous depth image $D_{t-1}$. To do this, we first back-project each of the two depth images via
\begin{equation}
\footnotesize
\Pi_i^{-1}(\mathbf{u}) \equiv \pi_{K,i}^{-1}(\mathbf{u}) = D_i(\mathbf{u})K^{-1}\dot{\mathbf{u}},
\end{equation}
in which $K$ denotes the camera intrinsics matrix and $\dot{\mathbf{u}} = (\mathbf{u}^\top, 1)^\top$, to make camera-space points images $\Pi_t^{-1}$ and $\Pi_{t-1}^{-1}$. We then transform these into world-space points images $W_t$ and $W_{t-1}$ using the known poses ${}_wT_t$ and ${}_wT_{t-1}$ of the two frames. Next, we construct a selection image $\Phi$ that establishes reprojection correspondences between the two frames, based on the estimated depths in frame $t$. Formally,
\begin{equation}
\footnotesize
\Phi(\mathbf{u}) = \rho(\pi_{K,t-1}({}_wT_{t-1}^{-1} * W_t(\mathbf{u}))),
\end{equation}
where $*$ denotes matrix multiplication, $\pi_{K,i}$ projects a point in the camera space of frame $i$ back down onto the image plane, and $\rho$ rounds a point in $\mathbb{R}^2$ to the nearest integer coordinates. Finally, we filter out any pixel $\mathbf{u}$ in $D_t$ that (i) has a correspondence $\Phi(\mathbf{u})$ that falls outside the bounds of $D_{t-1}$, or (ii) satisfies $\|W_t(\mathbf{u}) - W_{t-1}(\Phi(\mathbf{u}))\|_2 > 0.1$.

\emph{Depth Truncation:} Second, we filter out pixels with an estimated depth that is greater than 4 metres, since distant points are less likely to be reliably estimated. 

\emph{Small Depth Region Removal:} Third, like \cite{Pradeep2013}, we segment the depth image into components whose depth values differ by $\le$ a threshold, and remove components below a certain size. To implement this, we make a binary edge map
\begin{equation}
\footnotesize
E(\mathbf{u}) = \mathbb{I}\left[ \exists \mathbf{n} \in \mathcal{N}(\mathbf{u}) \mbox{ s.t. } \left\|D(\mathbf{n}) - D(\mathbf{u})\right\|_2 > \tau_d \right],
\end{equation}
in which $D$ denotes the depth image, $E$ denotes the edge map, $\mathbf{u}$ denotes a pixel in the domain of $D$ and $E$, $\mathbb{I}$ denotes the binary indicator function, $\mathcal{N}$ denotes the ($5 \times 5$) neighbourhood of $\mathbf{u}$, and $\tau_d = 0.05$. We find the connected components of $E$ and remove any whose size is $< 5000$.

\emph{Median Filtering:} Finally, we median filter the depth (with a $5 \times 5$ kernel) to produce a smoother reconstruction.

\subsection{Mapping Client}
\label{subsec:mappingclient}

\begin{figure}[!t]
	\vspace{2mm}
	\begin{center}
		\includegraphics[width=\linewidth]{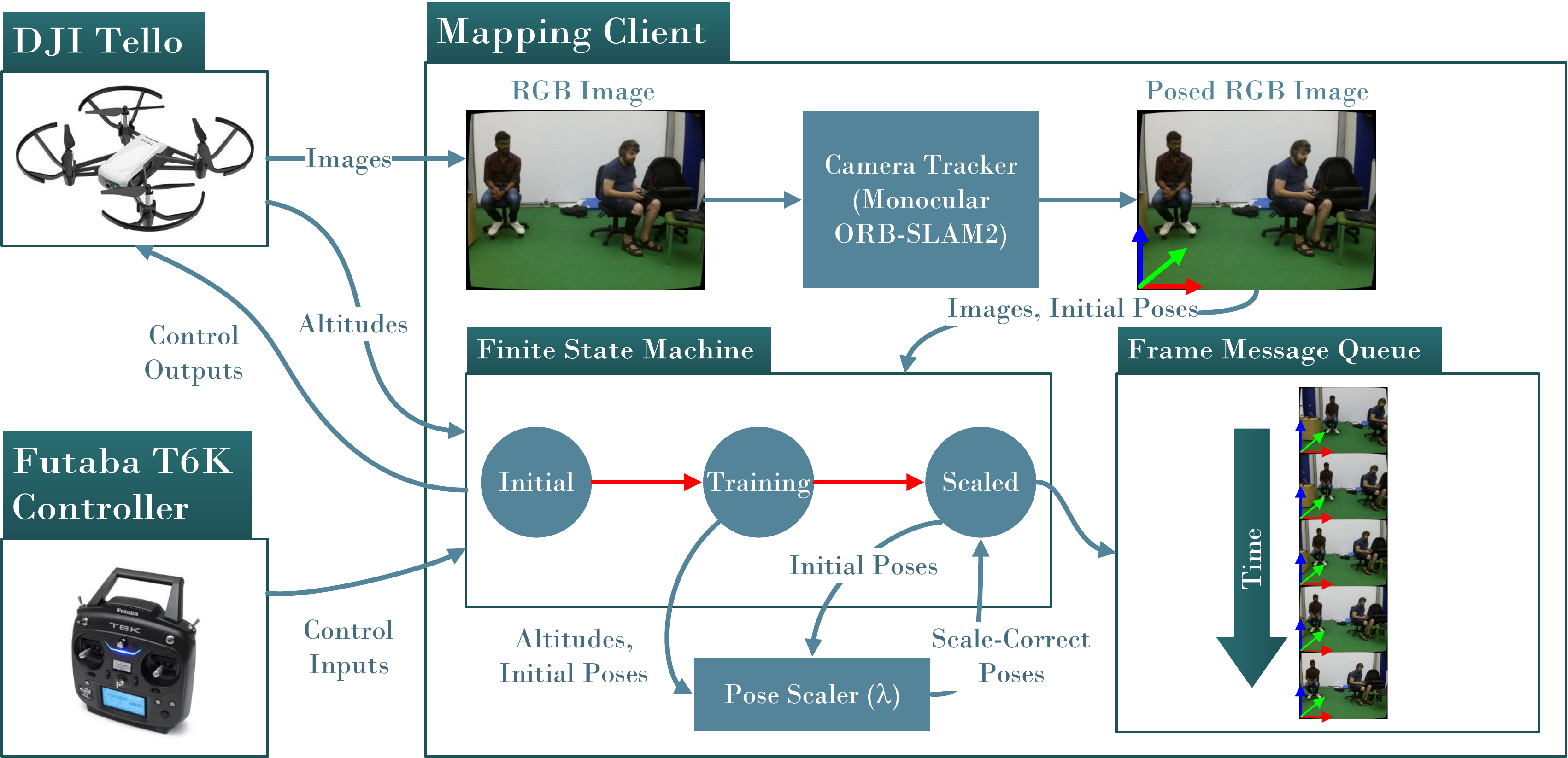}
		\caption{Our drone mapping client (\S\ref{subsec:mappingclient})}
		\label{fig:dronemappingclient}
	\end{center}
	\vspace{-1.5\baselineskip}
\end{figure}

Mapping clients send a sequence of RGB images and their 6D poses to the mapping server for use in updating the map. Various clients are possible. This section describes our drone mapping client, which we used to capture our new dataset. It requires a drone with a fairly accurate altitude sensor, e.g.\ the downward-facing time-of-flight sensor on the DJI Tello.

The client (see Figure~\ref{fig:dronemappingclient}) is based on a finite state machine, with states indicating the progress made in establishing scale-correct tracking. It starts in the \emph{initial} state, with the drone on the ground and the camera tracker (in practice, monocular ORB-SLAM2 \cite{MurArtal2017}) not yet having started. It stays in this state throughout the take-off. After the first take-off, the tracker is started, and begins to emit a non-scale-correct pose for each frame. The user can then fly around arbitrarily and/or manually trigger a transition to the \emph{training} state. In the \emph{training} state, the user is restricted to moving only up or down (since we don't know the orientation of the tracker's coordinate system with respect to the ground), and must execute a series of up and down motions to allow the client to estimate the correct scale. The estimation involves sampling, over the course of the training process, a sequence of scale-correct altitudes $a_1$, ..., $a_n$ and their corresponding tracker positions $\bm{p}_1$, ..., $\bm{p}_n$, and then computing the scale ($\lambda$) as
\begin{equation}
\footnotesize
\lambda = \frac{\sum_{i=1}^{n-1} |a_{i+1} - a_i|}{\sum_{i=1}^{n-1} \left\|\bm{p}_{i+1} - \bm{p}_i\right\|_2}
\end{equation}
The user manually concludes the training process by triggering a transition to the \emph{scaled} state. In this state, the user can again fly around normally, but poses emitted by the tracker are scaled using $\lambda$, and the images and their resulting scaled poses are streamed across to the mapping server.

\section{Experiments}
\label{sec:experiments}

\noindent We perform extensive experiments on three datasets:

\emph{ScanNet} \cite{Dai2017CVPR} is a large-scale 3D reconstruction dataset consisting of 1513 RGB-D sequences of unpopulated scenes. We use it to compare the different depth estimators we consider (\S\ref{subsec:experiments-scannet}), and to show that our post-processing (\S\ref{subsec:staticmapping}) significantly improves the accuracy of the reconstructions produced, for only a small cost in terms of completeness.

\emph{GTA-IM} \cite{Cao2020} is a large-scale indoor motion and human-scene interaction dataset consisting of 64 synthetic \mbox{RGB-D} sequences with ground-truth camera poses and instance masks, and 3D skeleton data for the primary character in each sequence. We use it to compare the skeleton detectors (\S\ref{subsubsec:experiments-gta-im-skeletons}) and people masking (\S\ref{subsubsec:experiments-gta-im-peoplemasking}) approaches we consider.

We also collect our own \emph{Oxford Hybrid Mapping} (OHM) dataset, which consists of 6 populated scenes, using an Asus ZenFone AR, a DJI Tello and a Vicon Vantage system. For each scene, we provide: (i) a ground-truth mesh reconstructed by running InfiniTAM \cite{Prisacariu2017} on an RGB-D sequence of the static scene, captured in advance with the ZenFone using TangoCapture \cite{Golodetz2018TVCG}; (ii) an RGB sequence of people then moving around the scene, captured from the DJI Tello and observed by the Vicon Vantage system; and (iii) the relative transformation between the coordinate system of the Tello's camera and the (arbitrarily placed) coordinate system of its Vicon subject. We place Vicon markers on the Tello to allow its trajectory to be tracked, and the people wear motion capture suits so that the Vicon system can track their skeletal trajectories. We use our dataset to evaluate the performance of the 3D reconstruction (\S\ref{subsubsec:experiments-ohm-reconstruction}) and skeleton detection (\S\ref{subsubsec:experiments-ohm-skeletons}) components of our hybrid mapping system on real data captured from a monocular drone.

\subsection{ScanNet}
\label{subsec:experiments-scannet}

\begin{table}[!t]
	\vspace{2mm}
	\centering
	\caption{Reconstruction results averaged over the ScanNet v2 test set. `PP' denotes post-processing. The full post-processing we perform has both a spatial and a temporal component (see \S\ref{subsec:staticmapping}). We report results both with and without this enabled. For DeepVideoMVS, we also report ablated results with only the spatial component of the post-processing, to show our temporal post-processing helps even for a method that already considers temporal consistency.}
	\label{tbl:scannet-reconstruction}
	\scriptsize
	\begin{tabular}{ccc}
		\textbf{Method} & \textbf{Mean Inaccuracy (m)} & \textbf{Mean Incompleteness (m)} \\
		\midrule
		MVDepthNet \cite{Wang20183DV} & 0.800 & 0.097 \\
		+ (full) PP & 0.275 & 0.106 \\
		\midrule
		DeepVideoMVS \cite{Duzceker2021} & 0.123 & \textbf{0.085} \\
		+ spatial PP (only) & 0.108 & 0.090 \\
		+ (full) PP & \textbf{0.079} & 0.112 \\
		\bottomrule
	\end{tabular}
	\vspace{-1.7\baselineskip}
\end{table}

To compare MVDepthNet \cite{Wang20183DV} and DeepVideoMVS \cite{Duzceker2021} in the context of our own post-processing of the estimated depth images, we use each, both with and without post-processing, to reconstruct TSDFs of the 100 RGB-D sequences in the ScanNet v2 test set \cite{Dai2017CVPR}, and compare the vertex clouds of meshes constructed from these TSDFs to the ground truth using the cloud-to-cloud distance metric from CloudCompare \cite{CloudCompare}. This (asymmetric) metric is defined as
\begin{equation}
\footnotesize
\mathit{C2C}(S, T) = \frac{1}{|S|} \sum_{\bm{s} \in S} \min_{\bm{t} \in T} \|\bm{t} - \bm{s}\|_2,
\end{equation}
where $S$ and $T$ denote the two point clouds. When $S$ is the vertex cloud of the estimated reconstruction and $T$ is that of the ground truth, this measures the inaccuracy of the reconstruction; the converse measures its incompleteness.

Table~\ref{tbl:scannet-reconstruction} shows the results. As expected, DeepVideoMVS consistently outperforms MVDepthNet, with or without our post-processing. More interestingly, our post-processing approach is consistently able to significantly improve the accuracy of the reconstructions produced by either method, with only a small corresponding drop in completeness. This is particularly noteworthy for DeepVideoMVS, which is based on a ConvLSTM and thus already expected to take temporal consistency into account to some extent. Moreover, a significant proportion of the increase in accuracy we achieve can be attributed to the world-space temporal filtering step of our approach, as shown by our ablated results in Table~\ref{tbl:scannet-reconstruction}.

Figure~\ref{fig:scannet-reconstruction} shows a visual example. DeepVideoMVS is much less noisy than MVDepthNet even prior to post-processing, but note that even with a modern method, significant accuracy improvements are possible for only a small completeness cost. Also note the major improvements that our post-processing achieves for an older method like MVDepthNet.

\begin{stusubfig}{!t}
	\vspace{2mm}
	\begin{subfigure}{.23\linewidth}
		\centering
		\includegraphics[width=\linewidth]{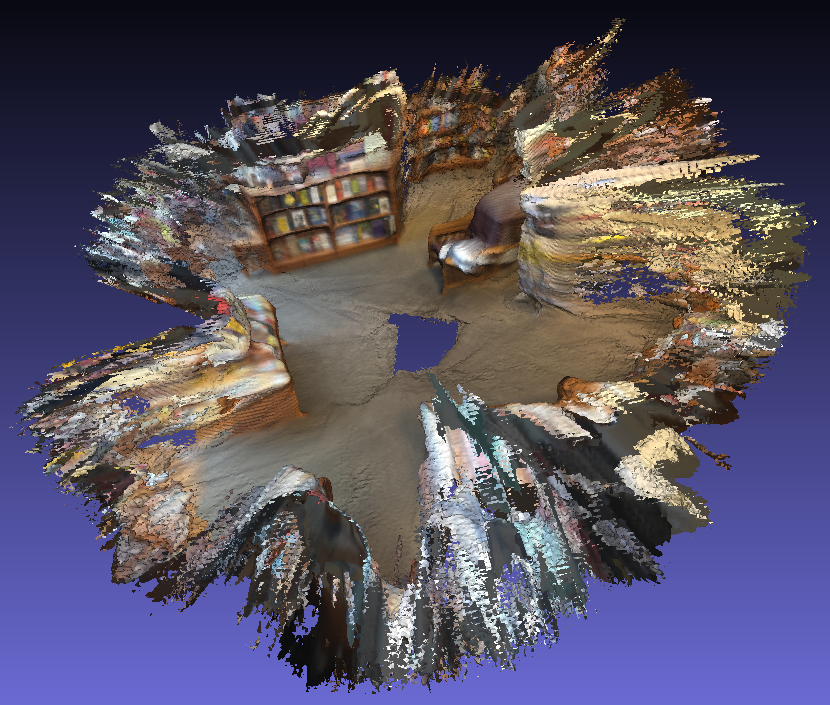}
		\caption{\scriptsize MVDepth}
	\end{subfigure}%
	\hspace{2mm}%
	\begin{subfigure}{.23\linewidth}
		\centering
		\includegraphics[width=\linewidth]{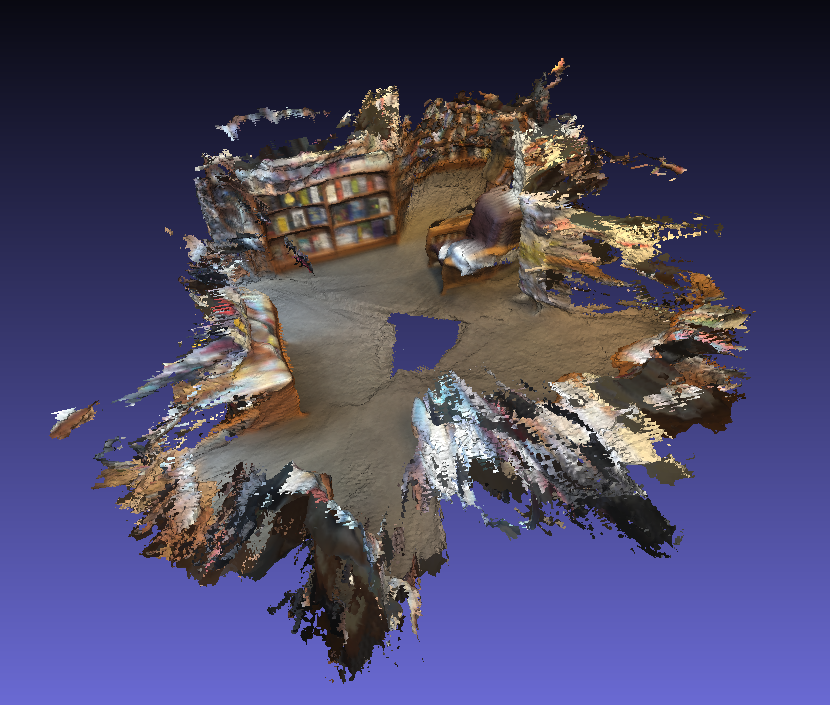}
		\caption{\scriptsize MVDepth+PP}
	\end{subfigure}%
	\hspace{2mm}%
	\begin{subfigure}{.23\linewidth}
		\centering
		\includegraphics[width=\linewidth]{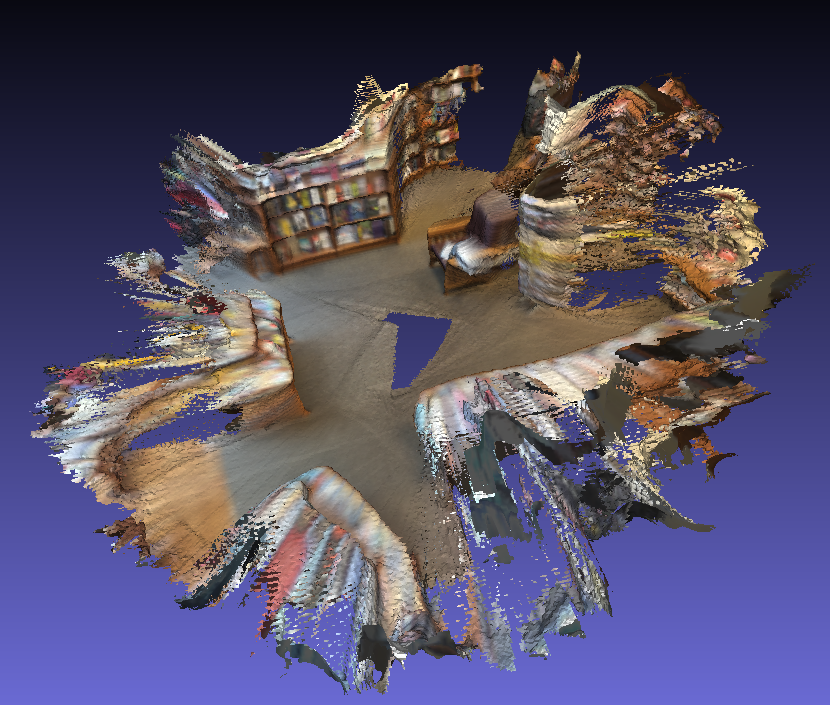}
		\caption{\scriptsize DVMVS}
	\end{subfigure}%
	\hspace{2mm}%
	\begin{subfigure}{.23\linewidth}
		\centering
		\includegraphics[width=\linewidth]{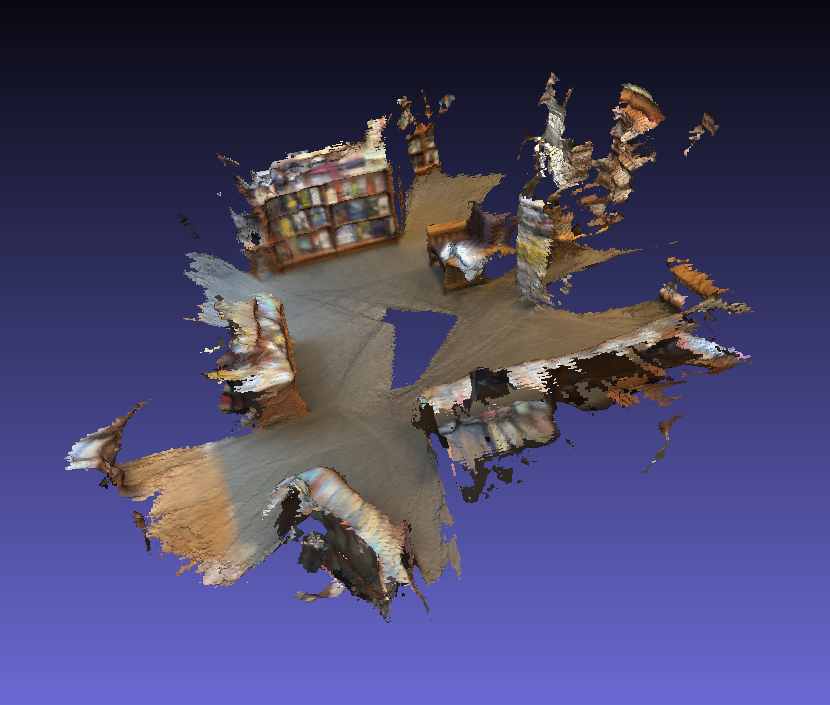}
		\caption{\scriptsize DVMVS+PP}
	\end{subfigure}%
	\caption{Comparing the reconstructions produced by MVDepthNet and DeepVideoMVS, with and without our post-processing, on \textit{scene0800\_00} from ScanNet.}
	\label{fig:scannet-reconstruction}
	\vspace{-1.5\baselineskip}
\end{stusubfig}

\subsection{GTA-IM}
\label{subsec:experiments-gta-im}

To compare the different skeleton detectors and people masking approaches we consider, we evaluate them on the 14 RGB-D sequences in the 5 Hz subset of GTA-IM~\cite{Cao2020}.

\subsubsection{Skeleton Detection}
\label{subsubsec:experiments-gta-im-skeletons}

We compare LCR-Net~\cite{Rogez2017CVPR} and XNect~\cite{Mehta2020TOG} using the well-known mean per-joint position error (MPJPE) and \% of correct 3D keypoints (3DPCK) metrics. Since our main interest is in performing multi-person 3D skeleton detection from a moving drone, it is important to investigate how both detectors perform in an environment like ours, rather than the static camera environment \cite{Mehta20183DV} in which they were compared in \cite{Mehta2020TOG}.

Table~\ref{tbl:gta-im-skeletons} shows the results. Notably, both achieve significantly worse results than they would in their normal contexts, for several reasons: (i) there is a domain transfer issue in testing models trained on images of real people on a synthetic dataset like GTA-IM; (ii) XNect makes temporal consistency assumptions that are not well designed for dealing with fast camera movements (and especially jumps in the camera trajectory); and (iii) GTA-IM consists of sequences in which the camera follows a person around in third-person view, often from a slightly downward-looking angle, and at times with the lower half of the person out of view. Nevertheless, it is clear that \mbox{LCR-Net} outperforms XNect in this context, and by a considerable margin, something that we also observed in our real-world drone experiments (\S\ref{subsubsec:experiments-ohm-skeletons}).

\subsubsection{People Masking}
\label{subsubsec:experiments-gta-im-peoplemasking}

As per \S\ref{subsubsec:peoplemasking}, we consider $5$ different ways of masking people out of the input images to prevent them from being fused into the map, namely LCR-Net + Bounding Volumes (BVs), LCR-Net + SMPL, XNect + BVs, XNect + SMPL, and Mask R-CNN \cite{He2017}. We compare the masks they produce to the ground-truth masks using the well-known intersection-over-union (IoU) and F1 metrics. We also evaluate the \emph{coverage ratio} (CR) of the masks, where $\mathit{CR}(M, M_G) = |M \cap M_G| / |M_G|$ is the coverage ratio achieved by a mask $M$ with respect to a ground-truth mask $M_G$. The coverage ratio measures the ability of a method to effectively mask people out of the image, which can differ from its ability to produce a precise segmentation of the people. We compute these three metrics for each individual frame, and then average them, firstly over all the frames in each sequence, and then over all the sequences we consider.

Table~\ref{tbl:peoplemasking} shows the results. The most effective method is clearly Mask R-CNN, but it is also much slower than the other approaches. All the other methods are much less accurate in matching the ground-truth masks, but both LCR-Net + BVs and LCR-Net + SMPL achieve relatively high CRs, whilst taking significantly less time than Mask R-CNN.

\begin{table}[!t]
	\vspace{2mm}
	\centering
	\caption{MPJPEs and 3DPCKs on the 5Hz subset of GTA-IM \cite{Cao2020}, averaged over all $14$ sequences.}
	\label{tbl:gta-im-skeletons}
	\begin{tabular}{ccccc}
		& \multicolumn{2}{c}{\textbf{MPJPE (m)}} & \multicolumn{2}{c}{\textbf{3DPCK@15cm (\%)}} \\
		& LCR-Net & XNect & LCR-Net & XNect \\
		\midrule
		\textbf{Average (all joints)} & 0.352 & 0.603 & 32.52 & 20.21 \\
		\bottomrule
	\end{tabular}
	\vspace{-1.5\baselineskip}
\end{table}

\subsection{Oxford Hybrid Mapping}
\label{subsec:experiments-ohm}

To evaluate our hybrid mapping system on real data from a monocular drone, we capture a dataset consisting of 6 populated scenes, as described previously.

\subsubsection{Populated Scene Reconstruction}
\label{subsubsec:experiments-ohm-reconstruction}

To evaluate the 3D reconstruction performance of DeepVideoMVS \cite{Duzceker2021} with our post-processing from \S\ref{subsubsec:postprocessing} on our dataset, we reconstruct a TSDF for each sequence in our dataset using Open3D, and compare the vertex cloud of the mesh constructed from each such TSDF to the vertex cloud of a ground-truth TSDF reconstructed using the system described in \cite{Golodetz2018TVCG} from an Asus ZenFone AR sequence of the same scene.

To compare these two reconstructions, they must be aligned, first in scale and then rigidly. First, we compute two lengths for the drone's trajectory: the first, $L_{vicon}$, is the ground-truth length according to the Vicon; the second, $L_{tracker}$, is the length according to the tracker, after online scale correction (\S\ref{subsec:mappingclient}). We scale the online tracker poses by $\frac{L_{vicon}}{L_{tracker}}$, a value that indicates the error in our online scale estimate for each sequence (see Table~\ref{tbl:ohm-results}). We then reconstruct the TSDF for the drone sequence using these corrected poses. Finally, we rigidly align the two correctly scaled reconstructions using the manual tools in CloudCompare \cite{CloudCompare}, and similarly use CloudCompare to calculate the mean inaccuracy of each reconstruction (see Table~\ref{tbl:ohm-results}).

The scales estimated online are within $13\%$ of the ground truth; we plan to improve this in future by using maximum-likelihood estimation \cite{Engel2014} and/or making use of the drone's IMU. The reconstruction inaccuracies are higher than with ScanNet (\S\ref{subsec:experiments-scannet}), since: (i) the parts of each scene captured by the drone and ZenFone sequences do not perfectly overlap, so even some accurately reconstructed points may not have a nearby ground-truth point; (ii) DeepVideoMVS \cite{Duzceker2021} was trained to perform well on ScanNet, but not on our dataset. However, Figure~\ref{fig:ohm-rigidalignment} shows good alignment between our reconstructions and the ground truth where they overlap.

\subsubsection{Skeleton Detection}
\label{subsubsec:experiments-ohm-skeletons}

We compare LCR-Net \cite{Rogez2017CVPR} and XNect \cite{Mehta2020TOG} on our dataset using the MPJPE and 3DPCK  metrics, as in \S\ref{subsubsec:experiments-gta-im-skeletons}. Here, we must first transform the detected skeletons into Vicon space for comparison with the ground-truth skeletons (the necessary relative transformation can be computed using the tracked trajectory of the drone's camera in Vicon space). A further complication is that ground-truth skeletons from the Vicon are available for every frame, whilst the corresponding people are only visible from the drone's camera for some frames, so we must avoid penalising the skeleton detectors for failing to detect people they cannot see. To do this, we manually specify that evaluation should be turned on/off for different Vicon subjects as they move in/out of view. Since multiple people may be present in a scene, we associate each ground truth skeleton with the closest detection (if any), and then average the MPJPE and 3DPCK metrics over all joints and all people in the scene.

\begin{table}[!t]
	\vspace{2mm}
	\centering
	\caption{People mask metrics for the 5 Hz subset of GTA-IM (\S\ref{subsubsec:experiments-gta-im-peoplemasking}). Notes: (1) CR = Coverage Ratio; (2) The times are for a single-person scene.}
	\label{tbl:peoplemasking}
	\scriptsize
	\begin{tabular}{ccccc}
		& \textbf{Mean IoU} & \textbf{Mean F1} & \textbf{Mean CR} & \textbf{Time (ms)} \\
		\midrule
		XNect + BVs    & 0.377 & 0.495 & 0.690 & 58 \\
		XNect + SMPL   & 0.436 & 0.544 & 0.713 & 67 \\
		LCR-Net + BVs  & 0.473 & 0.626 & 0.934 & 55 \\
		LCR-Net + SMPL & 0.563 & 0.699 & 0.919 & 64 \\
		Mask R-CNN     & \textbf{0.769} & \textbf{0.865} & \textbf{0.986} & 198 \\
		\bottomrule
	\end{tabular}
\end{table}

\begin{table}[!t]
	\centering
	\scriptsize
	\caption{The offline scale factors $\frac{L_{vicon}}{L_{tracker}}$, mean (rigid) reconstruction inaccuracies, MPJPEs and 3DPCKs for our system on our \emph{Oxford Hybrid Mapping} dataset. The skeleton metrics are averaged over all joints and all people.}
	\label{tbl:ohm-results}
	\begin{tabular}{ccccccc}
		\textbf{Scene} & \textbf{Scale} & \textbf{Mean Rec.} & \multicolumn{2}{c}{\textbf{MPJPE (m)}} & \multicolumn{2}{c}{\textbf{3DPCK@15cm (\%)}} \\
		(\textbf{\# People}) & \textbf{Factor} & \textbf{Inacc.\ (m)} & \cite{Rogez2017CVPR} & \cite{Mehta2020TOG} & \cite{Rogez2017CVPR} & \cite{Mehta2020TOG} \\
		\midrule
		S1 (1) & 1.07 & 0.190 & 0.200 & 0.491 & 49.75 & 7.11 \\
		S2 (1) & 0.93 & 0.206 & 0.148 & 0.478 & 59.44 & 2.21 \\
		S3 (1) & 0.92 & 0.225 & 0.177 & 0.552 & 45.26 & 0.88 \\
		M1 (2) & 0.87 & 0.322 & 0.200 & 0.534 & 36.78 & 6.51 \\
		M2 (2) & 1.04 & 0.279 & 0.457 & 1.106 & 13.88 & 2.71 \\
		M3 (3) & 1.11 & 0.267 & 0.298 & 0.656 & 29.69 & 4.11 \\
		\midrule
		\textbf{Avg.} & 0.99 & 0.248 & 0.247 & 0.636 & 39.13 & 3.92 \\
		\bottomrule
	\end{tabular}
	\vspace{-1.5\baselineskip}
\end{table}

Table~\ref{tbl:ohm-results} shows the results. As with GTA-IM (\S\ref{subsubsec:experiments-gta-im-skeletons}), LCR-Net outperforms XNect, due to the strong assumptions XNect makes about its target environment. In this case, the difference is even more pronounced, as the sequences in our dataset are less difficult than GTA-IM for LCR-Net (people's legs are in view much more of the time, and the frame-rate is higher) but more difficult for XNect (the drone flies below head height at times, and so the neck, which XNect relies on, can easily go out of shot as the drone approaches a person).

\section{Conclusion}
\label{sec:conclusion}

In this paper, we have presented what is, to our knowledge, the first system to perform simultaneous 3D reconstruction and multi-person 3D human pose estimation from a monocular camera mounted on a single UAV. Our system only requires three hardware components: a low-cost monocular drone (e.g.\ a DJI Tello), a GPU-enabled laptop, and a standard remote controller (e.g.\ a Futaba T6K). Furthermore, it can be deployed even in confined indoor spaces around people. We have validated the component-level design choices made for our system via extensive experiments on ScanNet \cite{Dai2017CVPR} and GTA-IM \cite{Cao2020}, and also constructed a new dataset, the \emph{Oxford Hybrid Mapping} dataset, to evaluate our system's overall performance. Both our system and our dataset will be made publicly available to help stimulate further work in this area. Moreover, given the loosely coupled nature of our system, we expect it to be relatively straightforward to introduce new methods for the different components of our system as they become available, allowing our system to function as a useful platform for further research over the longer-term.

\begin{stusubfig}{!t}
	\vspace{2mm}
	\begin{subfigure}{.47\linewidth}
		\centering
		\includegraphics[width=\linewidth]{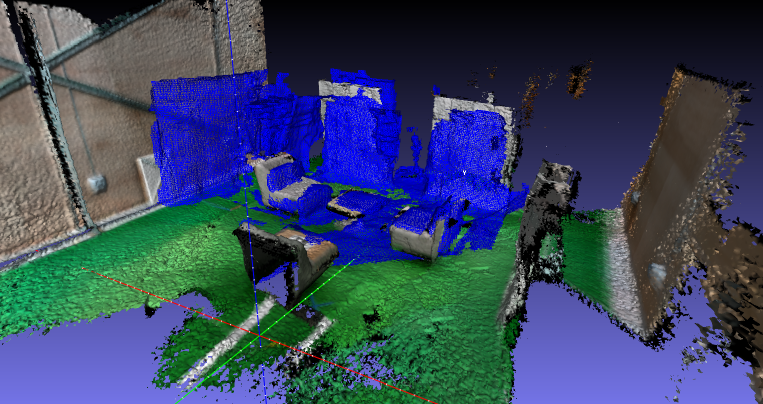}
	\end{subfigure}%
	\hspace{4mm}%
	\begin{subfigure}{.47\linewidth}
		\centering
		\includegraphics[width=\linewidth]{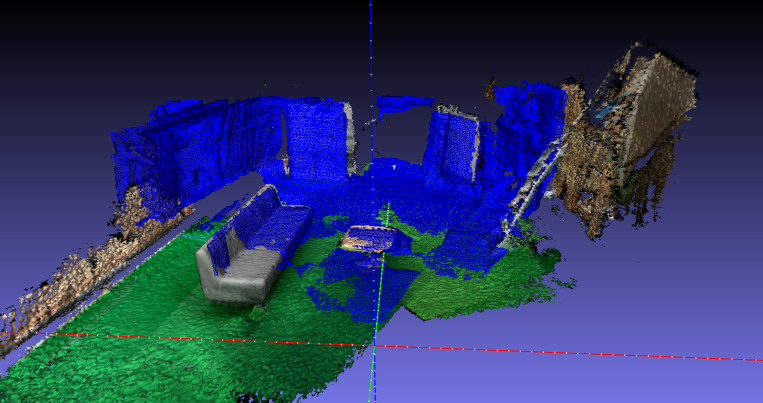}
	\end{subfigure}%
	\caption{Two examples (on scenes S1 and M2 of our dataset) of the rigid alignment between our reconstructions (blue) and the ground truth (coloured).}
	\label{fig:ohm-rigidalignment}
	\vspace{-1.5\baselineskip}
\end{stusubfig}

\section*{Acknowledgements}

This work was supported by Amazon Web Services via the Oxford-Singapore Human-Machine Collaboration Programme. We thank Graham Taylor for the use of the Wytham Flight Lab, Philip Torr for the use of an Asus ZenFone AR, and Tommaso Cavallari for implementing TangoCapture.

\bibliographystyle{IEEEtran}
\bibliography{hybridmappingpaper-arxiv}

\end{document}